\pdfoutput=1
\documentclass[11pt]{article}

\usepackage{naacl2021}
\usepackage{hyperref}

\usepackage{times}
\usepackage{latexsym}

\usepackage[T1]{fontenc}
\usepackage[utf8]{inputenc}

\usepackage{microtype}
\usepackage{graphicx}
\usepackage{enumitem}  % Added by tomerl@
\usepackage{array, multirow}  % Added by tomerl@
\usepackage{url}  % Added by tomerl@
\usepackage{amsmath}

\newcommand{\oidqe}[0]{Caption-Quality}
\newcommand{\captionext}[0]{Caption-Ext}
\newcommand{\ypred}[0]{\hat{y}}
\newcolumntype{H}{>{\setbox0=\hbox\bgroup}c<{\egroup}@{}}
\newcommand{\spearman}[0]{\rho^{\text{S}}}
\newcommand{\extgood}[0]{Ext-Good}
\newcommand{\modelqe}[0]{Q}

\title{Quality Estimation for Image Captions \\Based on Large-scale Human Evaluations}

\author{
  Tomer Levinboim~~~~~~~~Ashish V. Thapliyal~~~~~~~~Piyush Sharma~~~~~~~~Radu Soricut \\
   Google Research \\
   Venice, CA 90291 \\
  \texttt{\{tomerl,asht,piyushsharma,rsoricut\}@google.com}
}
\begin{document}
\maketitle

\begin{abstract}
  Automatic image captioning has improved significantly over the last few years, but the problem is far from being solved, with state of the art models still often producing low quality captions when used in the wild.
  %Usage of image captioning models in a production environment benefits from discarding poor quality captions. However, standard automatic metrics, such as CIDEr and SPICE,  cannot be used at prediction time since they require ground-truth references.
  In this paper, we focus on the task of Quality Estimation (QE) for image captions, which attempts to model the caption quality from a human perspective and \emph{without} access to ground-truth references, so that it can be applied at prediction time to detect low-quality captions produced on \emph{previously unseen images}.
  For this task, we develop a human evaluation process that collects coarse-grained caption annotations from crowdsourced users, which is then used to collect a large scale dataset spanning more than 600k caption quality ratings. We then carefully validate the quality of the collected ratings and establish baseline models for this new QE task.
 Finally, we further collect fine-grained caption quality annotations from trained raters, and use them to demonstrate that QE models trained over the coarse ratings can effectively detect and filter out low-quality image captions, thereby improving the user experience from captioning systems.
\end{abstract}

\section{Introduction}
Image captioning technology produces automatic image descriptions using natural language with the goal of being consumed by end-users that may not be able to directly access the images.
This need arises either because the user has a permanent condition (accessibility for visually impaired people), or due to a temporary situation where the user cannot use the visual modality (such as limited bandwidth, or smart voice-assistant).
In any of these situations, exposing the end-users to a generated caption that is incorrect negatively impacts user-trust, as it can have undesirable consequences for how they act next (for example, how they comment on a social-media site based on their misguided understanding).

In this paper, we propose to mitigate such risks  through Quality Estimation (QE) of image captions.
That is, we propose to automatically compute a quality estimation score $QE(image, caption)$ for a generated caption, and use it to control the quality of the captions presented to the user.  
For example, by filtering out captions with a low QE score (below a carefully chosen threshold), only high scoring captions would be served thereby minimizing the risks associated with low-quality captions.

We emphasize two aspects of QE that have guided us in our design choices:
First, the QE task is distinct from the model selection task: model selection measures output similarity to a fixed, ground-truth annotated dataset during training time (with traditional offline solutions such as CIDEr and SPICE). 
In contrast, a QE model estimates the caption quality with respect to the input image only and does so on \emph{previously unseen samples at prediction time} where ground-truth captions are unavailable.
Second, a QE model's goal is to assess the \emph{caption} as a whole and relate it to the \emph{image} content in a way that \emph{QE(image, caption)} aligns with human understanding of language and their perception of visual information.

To address these aspects we develop an image-caption evaluation process for collecting vast amounts of human judgements.
Specifically, we design the process to elicit only the type of human signal that is required for quality estimation -- human annotators are shown the image and asked to evaluate the caption as a whole by simply answering whether it is good or not. This type of high level feedback trades away the ability to understand in what way the caption is wrong, but its simplicity enables scaling up human evaluations to cover many more images, which promotes the generalization of the QE model to unseen images.

The dataset resulting from the evaluation process includes captions generated by various image-captioning model over 16,000 unique images from the Open Image Dataset~\cite{oidv4} for a total of ~55,000 \emph{unique} $\langle image, caption\rangle$ pairs, over which we collected approximately 600,000 binary human ratings.
We denote this dataset as \oidqe{}, provide extensive details on its generation process as well as make it publicly
available\footnote{\url{https://github.com/google-research-datasets/image-caption-quality-dataset}},
available.

The following summarizes our contributions:
\begin{enumerate}[topsep=0pt,itemsep=-1ex,partopsep=1ex,parsep=1ex]
\item We release the \oidqe{} dataset of roughly 65k human rated image-caption pairs, obtained by collecting approximately 600k binary human ratings in total. By analyzing the collected ratings, we show that they encode a stable and consistent signal about the caption.
\item We establish baseline results on the QE task and demonstrate that the signal encoded in the collected ratings is learnable, yet, cannot be trivially captured by an image-text similairty model trained over a large scale image-captioning dataset.
\item We further test our QE models, trained over the \oidqe{} dataset, and show that they can successfully rank correct-and-helpful captions higher than incorrect or unhelpful ones, even though they were never exposed to such a fine-grained signal. This is done by collecting additional fine-grained caption annotations from trained human raters, over images that are out-of-domain for the QE model.
\end{enumerate}
\section{Related Work}
Our paper is most similar to work done on evaluation metrics of image captions, where the main difference is that QE does not have access to the ground truth captions.
% Recent work in evaluation metrics for image captioning includes \cite{Cui2018LearningTE} which trains a classifier to distinguish between human and machine generated captions and applies it as an evaluation metric for model selection,
% as well as \cite{vifidel2019} which introduces VIFIDEL, a learned similarity function between the candidate description and object labels detected in the image.
% This similarity function assigns higher weights to object labels that appear frequently in reference captions.
% Interestingly, when tested for model selection, VIFIDEL with no references (a potential QE solution) correlates with human judgment almost as well as single reference BLEU or ROUGE.

Quality estimation has more than a decade long history in the Machine Translation (MT) field, from the early work based on feature engineering~\cite{specia2009estimating,soricut2010trustrank}, to more recent neural-network--based approaches~\cite{kreutzer2015quality,kim2016recurrent,kim2017predictorestimator}.
The QE track at the WMT conference~\cite{specia2018findings} has been running for several years, with multiple participants and notable improvements in model performance over the years.
However, there are significant differences in the formulation of the QE task between MT and image captioning, most notably the fact that the MT formulation is uni-modal (text-only alignment).
As a result, solutions for QE in the MT context tend to focus on feature-engineering that exploits this aspect~\cite{quest,kreutzer2015quality,martins2017pushing,wang2018alibaba}.
In contrast, QE for Image Captioning is a bi-modal problem (image-and-text alignment), and therefore better suited to approaches based primarily on deep feature representations and multi-modal feature integration, as we present in this paper.
%In particular, we show in this paper that our approach to QE for image captioning benefits from bi-modal pretraining of both image and text representations that are themselves produced by independent uni-modal pretrained models.

Beyond quality estimation modeling, the issue of effectively using quality estimators to improve the accessibility use-case for Blind or Visually Impaired (BVI) people has been previously studied~\cite{macleod2017understanding}.
The main question of their study is how to best inform the BVI user about the uncertainty around the generated captions, experimenting with framing the captions using phrases like ``I’m not really sure but I think it’s \$CAPTION'' or ``I’m 98\% sure that’s \$CAPTION''.
The findings are relevant in that BVI users of this technology have difficulties calibrating themselves into trusting or distrusting \$CAPTION, mostly because there is no alternative form of reference for the image content.
Therefore, if the caption provided to them (even accompanied by ``I’m not really sure but ...'') is in dissonance with the rest of the context (as it may be available in text form, e.g., as part of a tweet thread as in the study cited above),
they tend to resolve this dissonance not by believing that the caption is wrong, but by constructing scenarios or explanations that would somehow connect the two sources of information.
To mitigate this problem, we propose a thresholding-based approach that simply decides whether to show a caption or not based on a QE model's prediction (See section \ref{sec:extrinsic}).

\section{Building the \oidqe{} Dataset}
\label{sec:dataset}
The key contribution of this paper is the \oidqe{} dataset, a large collection of binary human judgments on the quality of machine-generated image captions (in English).
Below, we describe the dataset generation process, as well as the rating collection process with which we collect approximately 600,000 binary ratings via crowdsourcing. 
We then provide an analysis of the ratings which shows that they contain a consistent signal about the captions.
Note that in the experiments (section \ref{sec:extrinsic}), we further verify that indeed this signal captures the quality of the caption as perceived by \emph{trained} humans annotators.

\subsection{Image-Caption Generation}
\label{sec:caption_generation_models}
The starting point for our dataset is the Open Images Dataset (OID) \cite{oidv4} from which we randomly sample 16,000 images and then,
for legal and privacy concerns, filter out those which contain faces\footnote{Detected using the Google Cloud Vision API, \url{https://cloud.google.com/vision/}}.
The choice for OID images is driven by their image copyright status (CC BY) and the fact that they are out-of-domain for popular image captioning datasets such as COCO and Conceptual Captions.

To generate a diverse set of captions for annotation, we used several variants of Transformer-based \cite{vaswani2017attention} image-captioning models, trained on the Conceptual Captions dataset~\cite{sharma2018conceptual}, which consists of 3.3M training and $\sim$15,000 validation images-caption pairs.
As previous work indicates~\cite{sharma2018conceptual}, for out-of-domain images (OID), captions produced by Conceptual Captions trained models tend to have higher quality compared to captions produced by COCO-trained models.

All of the models are trained to minimize the ground-truth caption perplexity;
however, they differ on several important aspects (which contributes to caption diversity): the image feature representations, the number of object detection results they use, and the caption decoding procedure.
We briefly discuss these differences below; for further details, see
\cite{sharma2018conceptual,changpinyo2019decoupled}.

\paragraph{Global Image Representation}
Our captioning models use one of the following pretrained image encoders:
(1) The Inception-ResNet-v2 model \cite{szegedy2016inceptionv4}, %\marginpar{tomerl: Inception-ResNet-v2 - is this what INET uses?}
(2) The Picturebook image encoder~\cite{kiros2018illustrative}, or,
(3) The Graph-RISE model \cite{juan2019graphrise}, a ResNet-101 model~\cite{resnet2016} trained for an image classification task at ultra-fine granularity levels. % and graph-normalized to take advantage of image-similarity signals.

\paragraph{Object Representations}
The identification of objects in an image is done using a Faster R-CNN model, training it to predict both 1,600 object and 400 attribute labels in Visual Genome \cite{krishnavisualgenome},
following the Bottom-Up Top-Down setting \cite{anderson18bottomup}.
In terms of featurization for the identified bounding boxes, we use variants that include a ResNet-101 model pre-trained on ImageNet \cite{imagenet15} and one pre-trained using the Graph-RISE model \cite{juan2019graphrise}.

\paragraph{Object Labels} In addition to object-level representations, we detect object labels over the entire image, using a ResNet object-detection classifier trained on the JFT dataset~\cite{hinton2015distilling}.
The classifier produces a list of detected object-label identifiers, sorted in decreasing order by the classifier's confidence score.
These identifiers are then mapped to embeddings $o_j$ using an object-label embedding layer which is pre-trained to predict label co-occurrences in web documents using a word2vec approach~\cite{mikolov-et-al:2013a}.

\paragraph{Decoding} To further increase caption variance, we use either greedy decoding or beam search with beam width 5.

\begin{table}[b]
\center
\setlength\tabcolsep{4pt}
\begin{tabular}{lrrrr}
\textbf{Set} & \multicolumn{1}{l}{\textbf{Samples}} & \multicolumn{1}{l}{\textbf{\begin{tabular}[c]{@{}l@{}}Unique\\ Images\end{tabular}}} & \multicolumn{1}{l}{\textbf{\begin{tabular}[c]{@{}l@{}}Unique \\ Captions\end{tabular}}} & \multicolumn{1}{l}{\textbf{\begin{tabular}[c]{@{}l@{}}Unique\\ Models\end{tabular}}} \\ \hline
Train         & 58354                                & 11027                                                                                & 34532                                                                                   & 11                                                                                   \\
Dev           & 2392                                 & 654                                                                                  & 1832                                                                                    & 4                                                                                    \\
Test          & 4592                                 & 1237                                                                                 & 3359                                                                                    & 4
\end{tabular}
  \caption{The \oidqe{} dataset statistics}
\label{tbl:public_dataset}
\end{table}

\begin{figure}[t]
  \begin{center}
    \fbox{\begin{minipage}{0.85\linewidth}  % To remove the border, remove the \fbox command
      \includegraphics[width=1.0\linewidth]{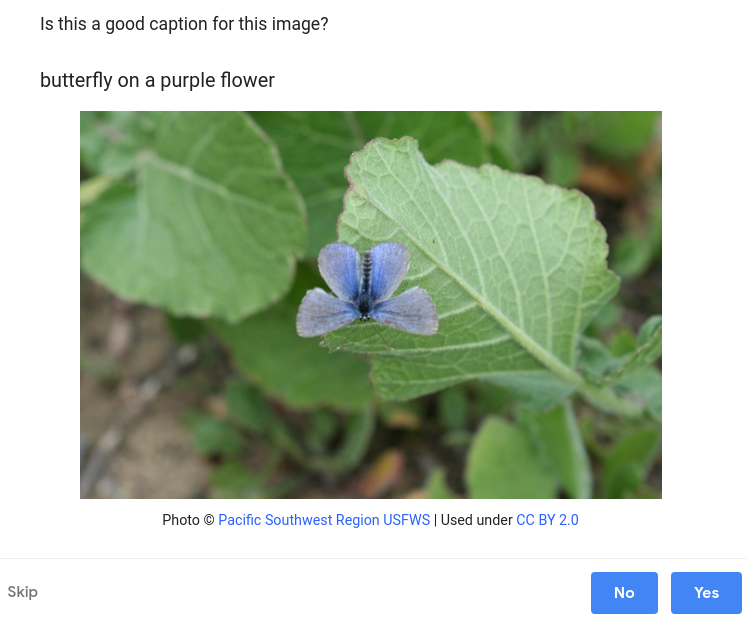}
    \end{minipage}}
  \end{center}
  \caption{
    Our caption evaluation interface.
    Raters indicate whether the caption is good/bad, or, they can skip.
  }
  \label{fig:interface_good_bad}
\end{figure}

\subsection{Fast\&Simple Human Annotation}
Traditional approaches for human evaluation of automatically generated text, such as for image captioning \cite{vinyals2016lessons} and machine translation \cite{banchs2015}, approach the task by collecting human ratings across multiple evaluation dimensions, such as correctness, informativeness and fluency. 
Such fine-grained evaluations are typically used to expose model deficiencies during development and can also assist during model selection. 
However, obtaining fine-grained rating on a large scale is a slow and costly process because it requires extensive manual labor by professionally trained human annotators. 
Furthermore, it is not immediately clear how the resulting multi-dimensional ratings can be combined to estimate the \emph{overall} caption quality in a human-like manner.

To avoid these complications we develop an evaluation process that asks the human evaluators to rate the generated text not per dimension, but \emph{as a whole}.
The benefits of our approach are threefold: 
(1) the collected ratings better align with our end goal of quality estimation from a human perspective
(2) having a single question accelerates caption evaluation, and 
(3) it substantially reduces the training and qualification requirements from the raters, which further contributes to the scalability of the evaluation process.

Specifically, we formulate the quality of an image-caption as the binomial probability $p=P(GOOD | image, caption)$ that can be estimated from the Bernoulli process in which every trial corresponds to a different rater.
We then leverage Google's crowdsourcing platform\footnote{\url{https://crowdsource.google.com}} on which we present (image, caption) pairs and ask \textbf{volunteer} raters the following coarse binary question,
$$\text{``\emph{Is this a good caption for the image?}''}.$$
The raters can then select YES/NO, or skip to the next sample (SKIP) (see Fig.~\ref{fig:interface_good_bad}).
In adopting this approach we take into account the fact that the platform's community consists of passionate volunteer raters, who may not have the linguistic background to provide fine-grained annotations. Furthermore, allowing the raters to skip captions reduces the risk of an undecided rater arbitrarily picking YES/NO just to move to the next image.

In order to reliably estimate the quality $p$ we collect a high number of $10$ ratings per image-caption sample. 
Once collected, the human ratings are further processed by:
(1) filtering out (image, caption) entries that received more than 2 SKIP ratings (practically, the vast majority of images were kept), and
(2) estimating $p$ by averaging the 8 to 10 ratings $r_i$ for each of the remaining (image, caption) pairs, and rounding to the closest score in $\{0, \frac{1}{8},\ldots,\frac{7}{8}, 1\}$, using the equation
$$\hat{p} = round(mean(r_i) * 8)/8,$$
where $r_i$ is 0 for NO answers and 1 for YES.

\begin{figure}[t]
  \begin{center}
    \includegraphics[width=1.0\linewidth]{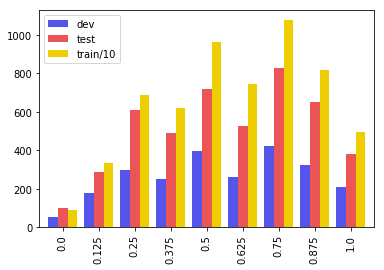}
  \end{center}
  \caption{
    A histogram of the dev, test and train $p^*$.
    The train set values were divided by 10 (for scale).
  }
  \label{fig:dev_rating_hist}
\end{figure}

The resulting dataset, which we call the \oidqe{} v1.0 dataset, is then split into three image-disjoint subsets, used as train, dev and test folds in our experiments.
We provide statistics for these subsets in Table~\ref{tbl:public_dataset}, as well as histograms of $\hat{p}$ in Fig.~\ref{fig:dev_rating_hist}.
Finally, we provide examples from the dev set in Table~\ref{tbl:oidqe-samples}.

\begin{figure}[b!]
  \begin{center}
    \includegraphics[width=1.0\linewidth]{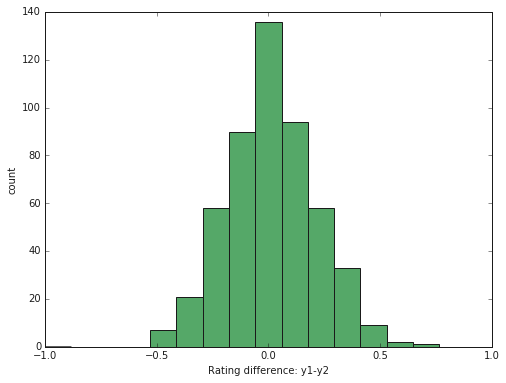}
  \end{center}
  \caption{
    A set of 509 captions were evaluated twice by different sets of 10 raters and 4 weeks apart.
    The figure shows a histogram of average human score differences $(\hat{p}_1 - \hat{p}_2) \in [-1,1]$,
    with scores $\hat{p}_i$ ($i\in\{1,2\}$) collected during the i-th evaluation.
    85\% of pairs are within a 0.25 distance, indicating that the evaluation setup produces a consistent and reliable signal.
  }
  \label{fig:rating_stability}
\end{figure}

\subsection{Stability Analysis}
\label{sec:dataset:stability}
As described above, the interpretation of what a ``GOOD'' caption means is left up to the raters, which could lead to unstable or inconsistent human ratings \cite{graham-etal-2013-crowd}.
In order to verify the stability of the quality ratings $\hat{p}$, we study the degree of agreement between different sets of 10 raters.
We ran an evaluation over the same set of 509 image-captions twice, but 4 weeks apart\footnote{The evaluation platform roughly guarantees that the ratings are provided by different subsets of raters.}.
An analysis of the difference of scores $(\hat{p}_1 - \hat{p}_2)$ over these 509 pairs results in an almost zero mean (mean=0.015) as well as low variance (std=0.212).
Figure~\ref{fig:rating_stability} provides a histogram of the differences $(\hat{p}_1 - \hat{p}_2)$ which clearly shows a concentration of the difference about 0.
Furthermore, repeating this analysis over a different set of image-captions results in similar statistics.

In conclusion, the stability analysis shows that by collecting and averaging 8-10 coarse binary ratings, we obtain consistent and reproducible P(GOOD) estimates $\hat{p}$  that are well-concentrated on a \emph{sample-level}.

\begin{table*}[t!]
\centering
\begin{tabular}{lll}
\textbf{Image} 
& \textbf{Generated captions} 
& \textbf{\begin{tabular}[c]{@{}l@{}}Human rating \end{tabular}} \\
\hline
\multirow{3}{*}{\includegraphics[width=0.14\linewidth]{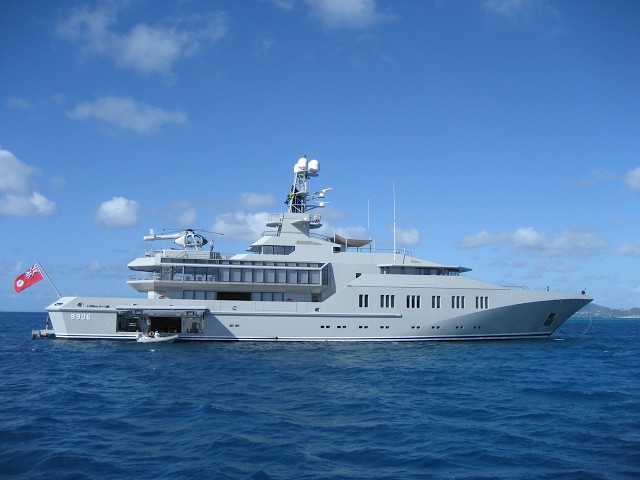}} 
& a general view of atmosphere . & 0.375\\
& this is a picture of a yacht . & 0.75 \\
&the yacht is a great place to take a rest . & 0.875\\
\hline

\multirow{3}{*}{\includegraphics[width=0.14\linewidth]{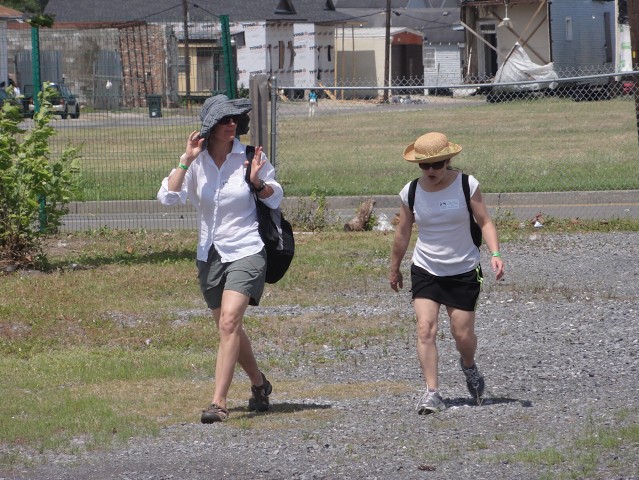}} 
& person and her husband take a walk . & 0.125\\
& people walking along the beach        & 0.25 \\
& people walking along the beach        & 0.5 \\

\hline
\multirow{3}{*}{\includegraphics[width=0.14\linewidth]{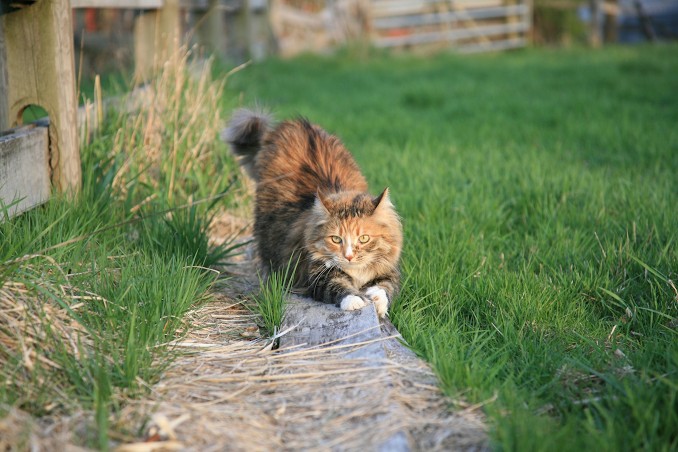}} 
& cat in the grass with a dog & 0.25\\
& a tiger in the grass          & 0.25 \\
& cat lying on the grass        & 0.75 \\
\hline

\multirow{3}{*}{\includegraphics[width=0.14\linewidth]{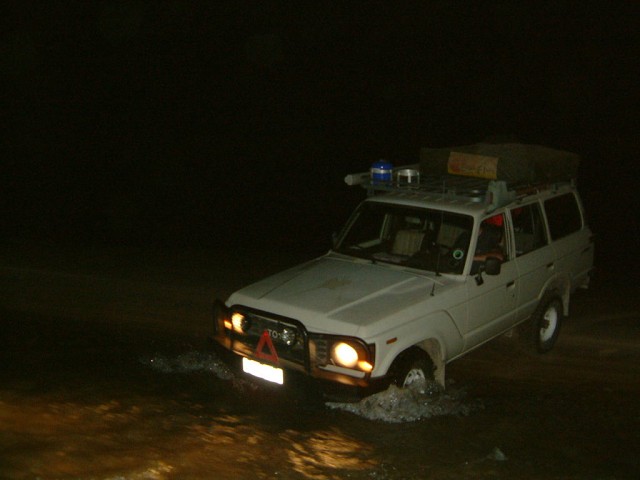}} 
& a police car in the middle of the road & 0.125\\
& automobile model in the rain . & 0.5 \\
& vehicles drive through a flooded street & 0.875 \\
\hline

\multirow{3}{*}{\includegraphics[width=0.14\linewidth]{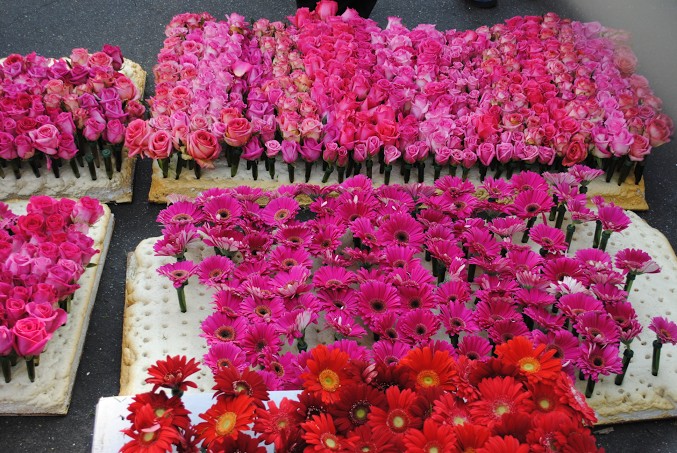}} 
& plants for sale at the local market & 0.625\\
& a selection of plants in the flower market & 0.875 \\
& flowers for sale at the market & 1.0 \\

\hline

\multirow{3}{*}{\includegraphics[width=0.14\linewidth]{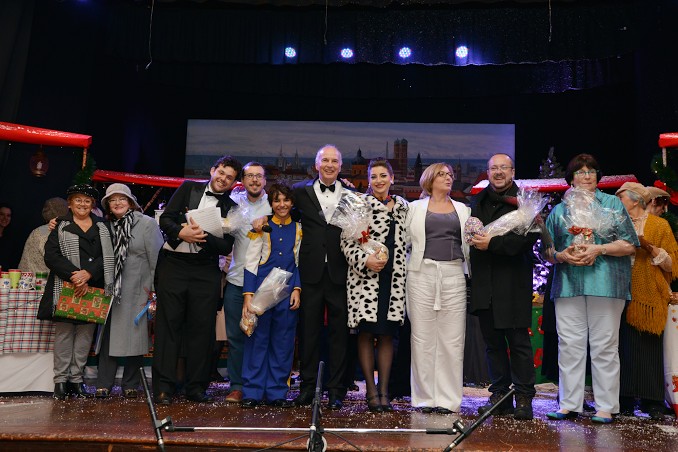}} 
& the team at the opening . & 0.375\\
& the cast performs on stage . & 0.5 \\
& the cast of musical film & 0.75 \\

% \multirow{3}{*}{\includegraphics[width=0.14\linewidth]{figs/dev_rated_captions/876e9f0b1dcfeb9c.jpeg}} & a young girl with a book and a cat on the table . & 0.0\\
%                                   & a photo of my daughter 's room . & 0.125 \\
%                                   & the table in the room        & 0.75 \\
% \\

\end{tabular}
  \caption{Samples from the \oidqe{} dataset  (dev fold). Images are paired with 3 captions and their corresponding mean human ratings.
  Repeated captions (which were generated by different captioning models) were rated by different sets of 10 raters, and tend to have similar scores (See stability analysis, cf. Figure~\ref{fig:rating_stability}).
  As can be seen, the higher scoring captions tend to include more information or contain fewer mistakes.
  }
\label{tbl:oidqe-samples}
\end{table*}

\begin{figure}[b!]
  \begin{center}
    \fbox{\begin{minipage}{0.95\linewidth}  % To remove the border, remove the \fbox command
      \includegraphics[width=1.0\linewidth]{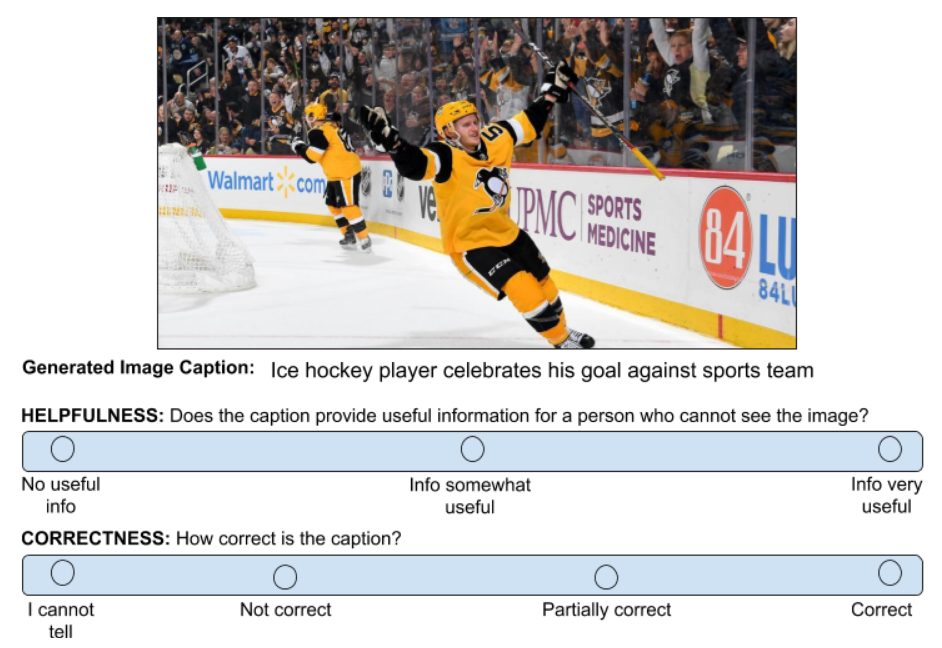}

    \end{minipage}}
  \end{center}
  \caption{
  Fine-grained evaluation interface presented to professional raters.
  The raters determine whether
  (1) the caption provides a helpful description for a person who cannot see the image,
  (2) the information in the caption is correct.
  }
  \label{fig:interace_fine_grained}
\end{figure}

\section{A Fine-Grained Caption Evaluation}
\label{sec:multi_dim}

We further collect fine-grained human annotations of image-captions to ascertain that the signal in the \oidqe{} dataset is beneficial for estimating the quality of image captions and filtering out low-quality ones.
Specifically, we ask professional human annotators to evaluate image-captions across two specific dimensions:
helpfulness and correctness\footnote{We also evaluate along a fluency dimension, but current captioning models tend to produce overall fluent outputs, which makes this dimension non-discriminative.}.
Fig~\ref{fig:interace_fine_grained} shows the evaluation interface.

Distinguishing between correctness and helpfulness is particularly crucial for quality estimation,
as it helps diagnose models that produce abstract or irrelevant captions which, while correct, do not provide useful image descriptions (specifically, for a person who is unable to see the image).
For example, consider the correct yet abstract caption ``Person in a sport event'' compared to the more descriptive caption
``Ice hockey player celebrates his goal against sports team'' (See Fig~\ref{fig:interace_fine_grained}). 
Another example of a correct but unhelpful caption is ``A view of the game from my living room'' because it conveys more information about the camera position rather than the actual image content.
While the previously discussed Fast\&Simple evaluation may assign all these captions with similar scores,
the fine-grained evaluation is capable of capturing such nuanced differences.

We posit that the large-scale annotations obtained by the Fast\&Simple approach will enable a model to distinguish between correct-and-helpful captions, and those that are not.
We ran the fine-grained evaluation once over 2,700 images, collecting 3 ratings per image.
The resulting dataset, denoted \captionext{} is used for our extrinsic QE evaluations (Sec. \ref{sec:extrinsic}).

\section{Models}
\label{sec:models}
This section presents a simple bilinear QE model which learns to combine the image and caption features to arrive at a quality estimate $QE(image, caption)$.
To construct the bilinear model we rely on expressive image and text representations that are produced by pretrained models that were themselves trained on vast amounts of uni-modal data.
Note that aside from building on top of pretrained models, we restrict further modeling to a simple architecture.
This was done in order to establish a baseline for our new QE task, as well as to remain focused on providing evidence that the signal in the $\oidqe$ dataset is both learnable and beneficial for quality estimation of image captions.
%When trained on the \oidqe{} dataset, this score reflects the model's prediction for how well $caption$ represents the aspects of $image$ that are important for the human raters.
% We then explore enhancing the QE model with bi-modal representations, learned by pretraining over image and text jointly.
% In particular, we contrast two types of bi-modal pretraining tasks
% classification vs generation (Sections \ref{sec:disc_pretrain}, \ref{sec:gen_pretrain}).

\subsection{A Bilinear QE model}
\label{sec:bilinear}

Our bilinear neural network model relies on three input types: caption, image and object labels.
These representations are produced by the following pretrained models:

\paragraph{Global Image Embedding}
For a global image representation, we used the latest Graph-RISE model version~\cite{juan2019graphrise} which produces a compact image embedding $i$ of dimension $D_i=64$.
Using this model enables transfer learning for QE with respect to image representation.

\paragraph{Object Labels Embeddings}
Objects present in the image (e.g. ``cat'', ``vehicle'', ``flower'') can help assess the correctness and helpfulness of a candidate caption,
where the intuition is that the caption should likely mention the more salient objects.
We use the object label model mentioned in Sec.~\ref{sec:caption_generation_models}, whose resulting embedding sequence is $O=(o_1, \ldots, o_{|O|})$, where each $o_j$ has dimension $D_o=256$.

\paragraph{Caption Universal Sentence Embedding}
The caption text is embedded using a pretrained version of the Universal Sentence Encoder (USE)~\cite{cer2018universal} into a $D_s=512$ dimensional vector $s$.
The USE model itself is trained on large amounts of English sources (Wikipedia, web news, discussion forums, etc.) and fine-tuned using supervised labels from the SNLI corpus~\cite{bowman2015large}.
We have alternatively tried a BERT~\cite{devlin2018bert} model as an encoder, but observed it provides no additional gains \cite{alikhani2020clue}
%Previous findings~\cite{conneau2017supervised} report an improvement in transfer-learning performance as a result of this setup.
\\

\begin{figure*}[t]
  \begin{center}
   \includegraphics[width=1.0\linewidth]{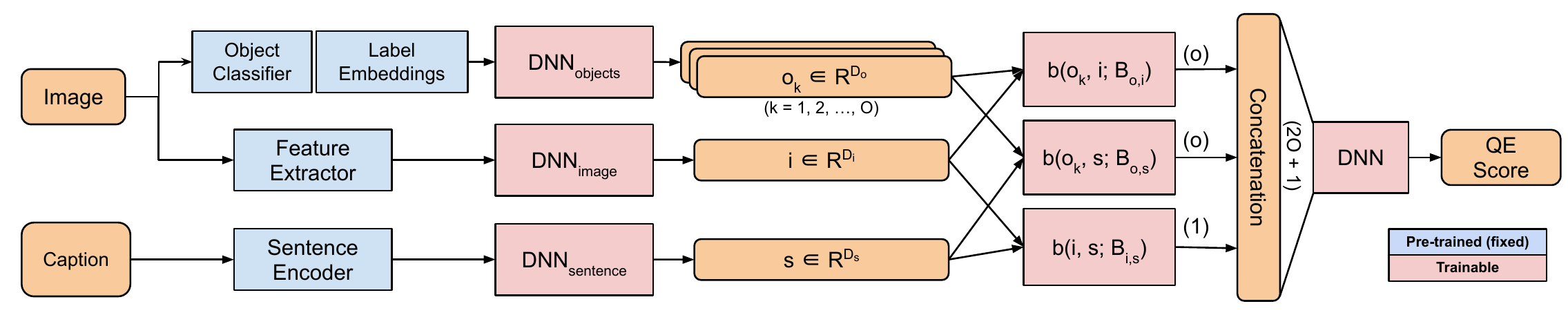}
   \caption{
     The bilinear QE model: Each input-modality pair has its own dedicated bilinear layer.
     The inputs to the model are pre-trained embeddings (blue) for the image, caption and object-label input types.
     The model parameters (pink) may further be warm-started by pretraining the model on an image-text similarity task (see Section \ref{sec:disc_pretrain}).
   }
  \label{fig:bilinear_model_full}
  \end{center}
\end{figure*}

Given these features, the bilinear QE model (illustrated in Figure~\ref{fig:bilinear_model_full}) processes each individual feature using a dense layer with a leaky-ReLU activation~\cite{xu2015empirical},
and then combines each of the resulting vector pairs using bilinear layers (see below).
All bilinear outputs are then concatenated and fed to a dense layer with a sigmoid activation, to produce the quality estimation $\ypred$.

\subsubsection{Bilinear Layers}
A bilinear layer models the inner product of its two inputs after applying a linear transformation to the second input.
This layer is defined as:
\begin{equation}
b(x, y; B) = x^{T}By = \langle x, By \rangle
\end{equation}
where $x \in R^{D_x}$ and $y \in R^{D_y}$ are input features, and $B \in R^{D_x \times D_y}$ is the learned parameter matrix.
Linear and bias terms can be added by appending a constant 1 to each of $x$ and $y$.

We use three such parameter matrices to capture the interaction between each pair of input-types:
\begin{enumerate}[topsep=0pt,itemsep=-1ex,partopsep=1ex,parsep=1ex]
\item $B_{o,i} \in R^{D_o \times D_i}$, applied to each of the object-label embeddings $[o_1, \ldots, o_{|O|}]$ and the image embedding $i$.
\item $B_{o,s} \in R^{D_o \times D_s}$, applied to each of the object-label embeddings $[o_1, \ldots, o_{|O|}]$ and the sentence embedding $s$
\item $B_{i,s} \in R^{D_i \times D_s}$, for the image embedding $i$ and sentence embedding $s$.
\end{enumerate}

\subsection{An Image-Text Similarity Baseline}
\label{sec:disc_pretrain}
Having the large scale Conceptual Captions dataset \cite{sharma2018conceptual} opens up the option to pretrain a QE model on an image-text similarity task \cite{Cui2018LearningTE} before fine-tuning on the \oidqe{} dataset.
We exercise this option by setting up a classification task whose goal is to match each image within a mini-batch with its corresponding ground truth caption.
Specifically, we feed the bilinear QE model mini-batches of size 256 and train it to detect the ground-truth caption of each image among the other ground-truth captions in the batch (along the lines of noise-contrastive estimation \cite{gutmann10a}).
The pretrained model achieves 62\% accuracy over the Conceptual Captions dev set and serves as an image-text similarity baseline.
In addition, its parameters serve as a fine-tuning initialization point that is better informed about the relationship between image and text compared to random initialization.

\begin{table*}[]
\begin{center}

\begin{tabular}{cl|cH|cc|ccH}
\textbf{\begin{tabular}[c]{@{}c@{}}\\Model\end{tabular}} & \textbf{\begin{tabular}[c]{@{}c@{}}\\QE training features\end{tabular}}  & \textbf{\begin{tabular}[c]{@{}c@{}}\\learning\\rate\end{tabular}}  & \textbf{\begin{tabular}[c]{@{}c@{}}LSTM\\ dim x layers\end{tabular}} & 
\textbf{ $\spearman_{\text{dev}}$} &
\textbf{$\spearman_{\text{test}}$} &
\textbf{$MSE_{\text{dev}}$} &
\textbf{$MSE_{\text{test}}$} &
\textbf{\begin{tabular}[c]{@{}c@{}}\\ AUC\end{tabular}} \\ 
\hline
Bilinear                                                                 & image, caption      & 1e-5 & N/A & 0.49  & 0.47 & 0.055 & 0.056 & 0.80 \\
Bilinear                                                                 & + 20 object labels  & 1e-5 & N/A & 0.50  & 0.47 & 0.055 & 0.058 & 0.82 \\
\hline
Bilinear (Pretrained)      
     & -                   & 1e-5 & N/A & 0.26  & 0.25 & 0.075 & 0.073 & 0.76                                                                \\
Bilinear (Pretrained)                                               & image, caption, 16 labels& 1e-5& N/A & \textbf{0.57}                                                             & \textbf{0.53}
                                                & \textbf{0.053} & \textbf{0.053}
                                                & \textbf{0.84}                                                   \\
% \hline
% LSTM (Gen')                                                          & $\gm{}$'s \emph{logits}                                               & 1e-6        & 2048x1                                                               & 0.47                                                            & 0.45                                                            & 0.76                                                            \\
% LSTM (Gen')                                                          & + \emph{seqlogp, sumlogp, seqlen}                                     & 1e-6        & 2048x3                                                               & 0.49                                                            & 0.47                                                            & 0.78                                                            \\
% \hline
% Combined (Gen')                                                      & (1) + $\gm{}$'s \emph{seqlogp}                                        & 1e-5        & 128x3                                                                & 0.54                                                            & 0.52                                                            & 0.78
\end{tabular}
\caption{Spearman's $\spearman$ scores on the Caption-Quality dev and test dataset (higher is better).
The pretrained and fine-tuned bilinear model exhibits the best Spearman results on the QE task. MSE results show the same trend and are included for completeness.
}
\label{tbl:exp_results}
\end{center}
\end{table*}

\section{Experimental Results}
\label{sec:results}
All QE models are trained on the \oidqe{} training set (Section~\ref{sec:dataset}).
We use Mean Squared Error ($MSE = \sum_{j=1}^{B}\frac{1}{N}(y_j-\ypred_j)^2$) as the loss function, where
$\ypred_j$ are the predicted scores and $y_j$ the ground-truth human scores.
For optimization, we use Adam~\cite{Kingma2015AdamAM} with batch size $B=256$ and tune the learning rate $lr\in\{1e\text{-}4,1e\text{-}5,1e\text{-}6\}$.
Dropout rate is set to $0.2$, and applied on the inputs of all trainable layers.
The following pretrained models are fixed during optimization:
%the image-captioning sub-module $\gm{}$,
the image encoder, the USE caption encoder, and object-label encoder.
The number of object-labels is tuned over $\{0, 5, 10, 20\}$, while the pretrained variants were fixed to 16.
%The stacked LSTM dimension$\times$layers is tuned over $\{128, 256, 1024, 2048, 4096\} \times\{1,2,3\}$.

Model selection is done by picking the checkpoint that maximizes the dev set Spearman's correlation $\spearman(y,\ypred)$.
Specifically, compared to MSE (the objective), the Spearman-based selection criterion better matches the intended use of the QE model, where at inference time, only images whose QE scores pass some threshold will be served. Since this threshold can be tuned, the absolute value of the predicted scores $\ypred$ is not as critical as obtaining a monotonic relationship between the predicted and ground truth scores (using $\spearman$ as the loss function is less feasible due to non-differentiability).

\subsection{Spearman's $\rho$ Analysis}
\label{sec:intrinsic}
We present in Table~\ref{tbl:exp_results} our dev and test Spearman results based on selecting the best-performing model configurations over the dev set.

Rows 1 and 2 show the bilinear model achieves minor improvements given additional 20 object labels.
The poor Spearman scores in row 3, which were obtained \emph{without} fine tuning over the \oidqe{} dataset, demonstrate that predicting the human ratings cannot be trivially achieved with an image-text similarity model, even when trained on a large dataset as Conceptual Captions.
On the other hand, after fine-tuning it for the QE task (row 4), both dev and test Spearman scores increase substantially by 6-7 Spearman points over the best non-pretrained variant, which demonstrates the effectivenss of bi-modal pretraining for the QE task.

% The best model among the stand-alone LSTM models (rows 5,6) is only able to match the performance of the non-pretrained bilinear model.
% This is despite using internal bi-modal representations of $\gm{}$, which was pretrained for caption generation.
% However, this result seems less surprising considering that $\gm{}$ learned to represent text by training over Conceptual Captions only, whereas the bilinear model variants invokes the USE encoder, which was trained over about two orders of magnitude more data~\cite{cer2018universal}.

% Row 7 shows that combining the two models (without pretraining for classification) improves upon the LSTM Spearman performance by around 3 Spearman points, a gain which we attribute to the two sub-modules being trained on information-complementary features.
% At the same time, even though the combined model has access to the USE caption representations, it is unable to match the best bilinear model (row 4), which points to bi-modal pretraining for classification being the missing ingredient.
% This implies that performance on the downstream task (QE) benefits more when the model is pretrained on a task with similar characteristics.

\subsection{Extrinsic Evaluation}
\label{sec:extrinsic}

% % Please add the following required packages to your document preamble:
% % \usepackage{graphicx}
% \begin{table}[b]
% \begin{tabular}{l|c|c}
% \begin{tabular}[l]{@{}c@{}}Model\\ category\end{tabular} & Features                                                                 & AUC   \\ \hline
% Confidence (2x2)                                         & \begin{tabular}[c]{@{}c@{}}logits+seqlogp\\ +sumlogp+seqlen\end{tabular} & 0.744 \\ \hline
% Confidence (2048x1)                                      & logits                                                                   & 0.755 \\ \hline
% Bilinear                                                 & 0 object-labels                                                          & 0.800 \\ \hline
% Bilinear                                                 & 20 object-labels                                                         & 0.818 \\ \hline
% Combined (2x2)                                           & ALL                                                                      & 0.804
% \end{tabular}%
% \caption{AUC scores for the best model per category.}
% \label{tbl:AUC}
% \end{table}

So far we have shown that the signal in \oidqe{} is both consistent and learnable.
In this section, we further show that the collected signal is effective for filtering out low-quality image captions.
To do so, we evaluate the performance of \oidqe{} trained QE models over the \captionext{} dataset, a more challenging setting which contains out-of-domain images (non-OID) and where each caption is annotated by three trained raters for its correctness and helpfulness (Sec. \ref{sec:multi_dim}).
Our analysis reveals that QE models trained over the \oidqe{} dataset generalize well to this harder task, having the ability to distinguish between correct-and-helpful image-captions and those that are not, even though these models were never exposed to such fine-grained signal.

\begin{figure}[b!]
  \begin{center}
    \fbox{
    \begin{minipage}{1.0\linewidth}  % To remove the border, remove the \fbox command
      \includegraphics[width=1.0\linewidth]{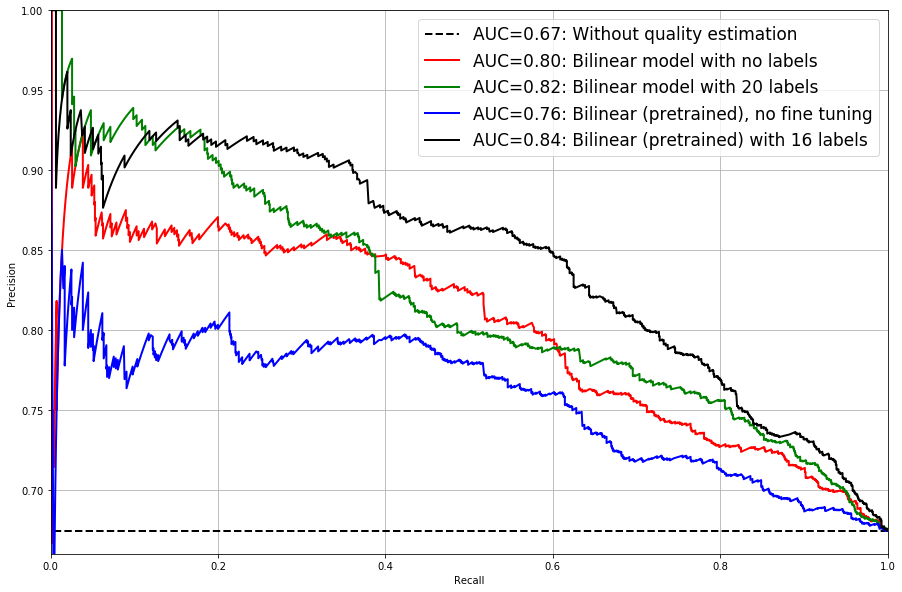}
    \end{minipage}
    }
  \end{center}
  \caption{
  Precision-Recall curves for the various Bilinear models. AUC values are reported in the legend.
  The pretrained and fine-tuned model (black) attains the highest precision values across almost all recall values.
  }
  \label{fig:precision_recall}
\end{figure}

Specifically, for a given image, we define a caption as \emph{\extgood{}} (extrinsically good) if a majority of raters agreed that it is at least partially-correct, and, a majority of raters agreed it is at least somewhat-useful.
With this definition, we  compute the \extgood{} precision and recall statistics of a QE model $\modelqe{}$ for each threshold $th\in[0,1]$ using the following equations:
\begin{equation}
precision^{\modelqe}_{th} = \frac{\sum_{s} 1^s_{\extgood} \cdot 1_{QE(s)>th}}{\sum_{s} 1_{QE(s)>th}}
\end{equation}
\begin{equation}
recall^{\modelqe}_{th} = \frac{\sum_{s} 1^s_{\extgood} \cdot 1_{QE(s)>th}}{\sum_{s} 1^s_{\extgood}}
\end{equation}
where the indicator variable $1^s_{\extgood}$ is on only when  $s$ is \extgood{}, and similarly the indicator variable $1_{QE(s)>th}$ is on only when the QE score of sample $s$ is higher than the threshold $th$.

Figure \ref{fig:precision_recall} shows the precision-recall curves and AUC scores for the same models analyzed in the previous section.
A visual inspection of this figure shows that the precision of the pretrained and fine-tuned bilinear model (black) dominates the other models across almost all recall values.
Indeed, in terms of AUC, the worst performing model is the image-text similarity baseline (blue; AUC=0.76) which has no access to the \oidqe{} dataset and its human ratings.
On the other hand, the pretrained and fine-tuned model (which is also the Spearman maxmizing model) attains the highest AUC score (AUC=0.84).

Put differently, to achieve precision=0.8 (i.e., 80\% of served captions are both correct and helpful), the image-text similarity model would be thresholded to serve only its top 21\% scoring image-captions (recall=0.21) while the pretrained and fine-tuned model would serve its top 71\% scoring image-captions (recall=0.71, or x3.4 improvement).
This analysis clearly demonstrates the usefulness of the \oidqe{} dataset for filtering out image-captions of low quality (where quality is determined by professional human raters).

% \subsection{\oidqe{} Dataset Usefulness}

% Furthermore, the \oidqe{} dataset and its ratings have already found usage beyond QE.
% In recent work \cite{seo2019reinforcing}, the \oidqe{} dataset was used to train an image-captioning model that is fine-tuned using off-policy reinforcement learning that uses the caption-quality annotations as the RL rewards.
% Their model achieves higher performance on the image-captioning task, as measured by human raters.
% This positive result demonstrates that the signal from our \oidqe{} dataset can come full-circle, providing useful auxilary labels for training better performing image-captioning models.

\section{Future Work}
\label{sec:future_work}
Beyond its relevance for the QE task, we expect that the collected signal in the \oidqe{} dataset will find usage in other image captioning tasks, such as (1) fine-grained caption evaluation (that is, caption classifiers that evaluate captions across multiple dimensions) for example, by way of pretraining against our dataset, as well as (2) improving caption generation itself, for example, by means of QE-based caption re-ranking, or by using the ratings in a reinforcement learning setup, as has recently been done by \cite{seo2019reinforcing}.

\section{Conclusion}
\label{sec:discussion}
In this paper we discussed how low-quality image-captions can negatively impact end-users and proposed a thresholding solution that relies on quality estimation of image captions, where caption quality is defined from a human perspective.
To make this solution feasible we developed a scalable human evaluation process with which we annotated a large number of image-captions with their human estimated quality.
We provided supporting evidence that the resulting dataset contains a consistent and reliable signal, as well as reported experimental results over professionally labeled fine-grained caption annotations, which verify that QE models trained over the \oidqe{} dataset are effective at filtering out low-quality image captions.

To encourage further research in automatic evaluation of image-captions, we make available our large-scale dataset of human judgments at \url{https://github.com/google-research-datasets/Image-Caption-Quality-Dataset}.

% Entries for the entire Anthology, followed by custom entries
\bibliography{main.bib}
\bibliographystyle{acl_natbib}

\end{document}